\pdfoutput=1
\documentclass[final,5p,times,twocolumn]{elsarticle}
\usepackage{amsmath,amsfonts}
\usepackage{algorithmic}
\usepackage{algorithm}
\usepackage{array}
\usepackage{textcomp}
\usepackage{stfloats}
\usepackage{url}
\usepackage{verbatim}
\usepackage{graphicx}
\usepackage{amssymb}
\usepackage{amsmath}
\usepackage{multirow}
\usepackage{graphicx}
\usepackage{booktabs}
\usepackage{bbding}
\usepackage[colorlinks,
            linkcolor=green,
            anchorcolor=green,
            citecolor=green]{hyperref}

\usepackage{amssymb}

\begin{document}

\begin{frontmatter}



\title{Textual Prompt Guided Image Restoration}

\author[1]{Qiuhai Yan}
\ead{yanqiuhai16@163.com}

\author[12,1]{Aiwen Jiang\corref{cor1}}
\ead{jiangaiwen@jxnu.edu.cn}

\author[1]{Kang Chen} 

\author[2]{Long Peng} 

\author[3]{Qiaosi Yi} 

\author[4]{Chunjie Zhang} 


\address[1]{School of Computer and Information Engineering, Jiangxi Normal University, Nanchang 330022, China}
\address[12]{School of Digital Industry, Jiangxi Normal University, Shangrao 333400, China}
\address[2]{University of Science and Technology of China, Hefei 230026, China}
\address[3]{School of Computer Science and Technology, East China Normal University, Shanghai 200062, China}

\affiliation[4]{organization={School of Computer and Information Technology, Beijing Jiaotong University},
            addressline={No.3 Shangyuancun},
            city={Haidian District},
            postcode={100044},
            state={Beijing},
            country={China}}

\cortext[cor1]{Corresponding author.}

\begin{abstract}
Image restoration has always been a cutting-edge topic in the academic and industrial fields of computer vision. Since degradation signals are often random and diverse, "all-in-one" models that can do blind image restoration have been concerned in recent years. Early works require training specialized headers and tails to handle each degradation of concern, which are manually cumbersome. Recent works focus on learning visual prompts from data distribution to identify degradation type. However, the prompts employed in most of models are non-text, lacking sufficient emphasis on the importance of human-in-the-loop. In this paper, an effective textual prompt guided image restoration model has been proposed. In this model, task-specific BERT is fine-tuned to accurately understand user's instructions and generating textual prompt guidance. Depth-wise multi-head transposed attentions and gated convolution modules are designed to bridge the gap between textual prompts and visual features. The proposed model has innovatively introduced semantic prompts into low-level visual domain. It highlights the potential to provide a natural, precise, and controllable way to perform image restoration tasks. Extensive experiments have been done on public denoising, dehazing and deraining datasets. The experiment results demonstrate that, compared with popular state-of-the-art methods, the proposed model can obtain much more superior performance, achieving accurate recognition and removal of degradation without increasing model's complexity. Related source codes and data will be publicly available on github site \href{https://github.com/MoTong-AI-studio/TextPromptIR}{https://github.com/MoTong-AI-studio/TextPromptIR}. 
\end{abstract}




\end{frontmatter}

\section{Introduction}
In practice, since weather(haze, rain, snow etc.) and recording medium are imperfect, images captured fail to represent scene adequately. Many degradation like distortion, blur, low contrast, color fading, noise etc. are introduced. As a prerequisite for intelligent applications, visually clear images are crucial for the success of downstream high-level tasks. Restoring visually pleasing high-quality images from given degraded ones, has therefore always been a cutting-edge topic in the academic and industrial fields of computer vision.

With development of deep learning, significant breakthroughs have been achieved in the field of image restoration. Traditionally, most of methods are trained for specific tasks, such as denoising\cite{torun2024hyperspectral,tian2023multi,wang2022blind2unblind,zhao2022hybrid}, deraining\cite{ozdenizci2023restoring,chen2023learning,peng2021ensemble,peng2020cumulative,wang2022self,yi2021structure,yan2023cascaded}, and dehazing\cite{yi2021efficient,xiao2024single,kumari2024new,del2022new}, and cannot cope with different types of degraded images in a single unified model. 

\begin{figure*}[h!]
\centering
\includegraphics[width=\textwidth]{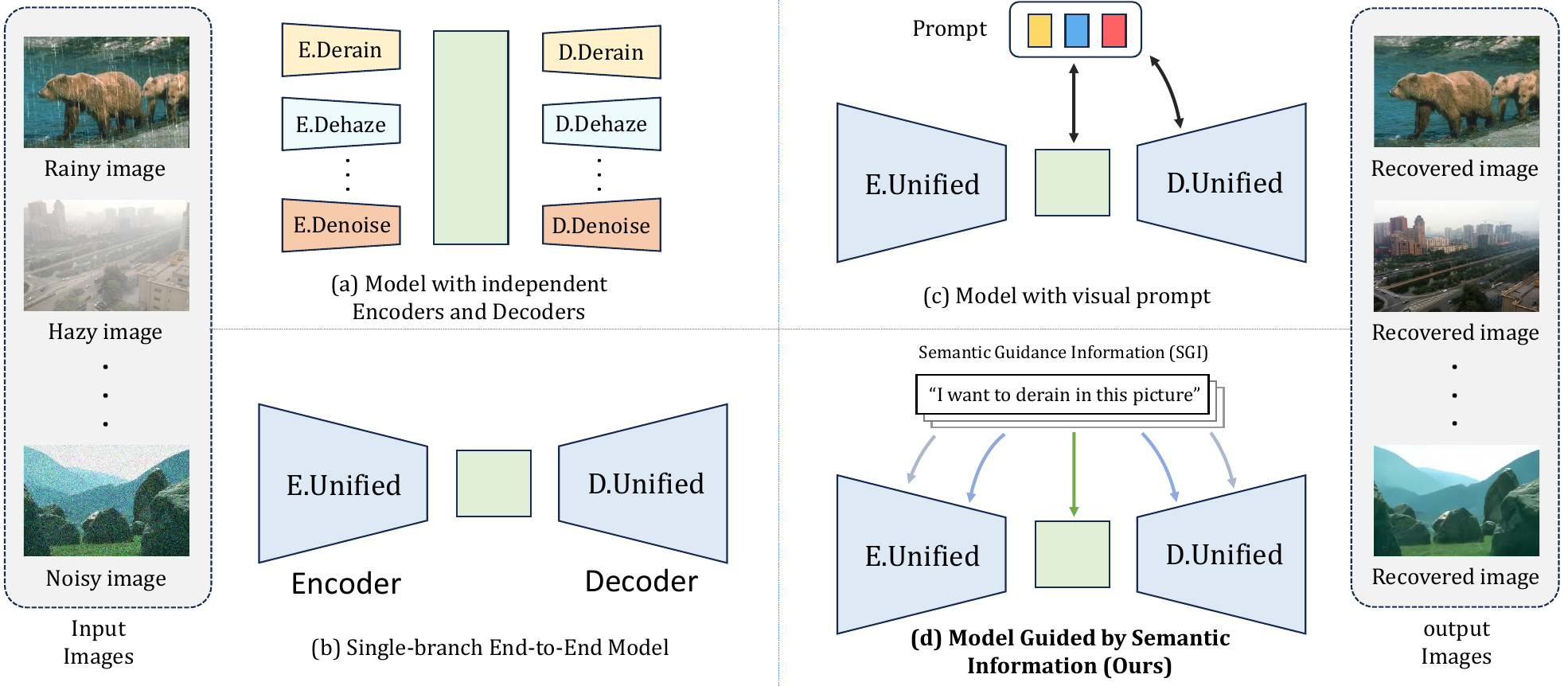}
\caption{Four categories of all-in-one image restoration methods. (a) model with independent Encoders and Decoders. (b) single branch end-to-end model. (c) model with visual prompt. (d) model with textual prompt (Ours), utilizing user-provided semantic guidance for image restoration.}
\label{fig_1}
\end{figure*}

Nevertheless, in the real world, the degradation signals are often random and diverse. "All-in-one" models that can do blind image restoration have been concerned in recent years. In Fig~\ref{fig_1}, four typical categories of "all-in-one" models are over-viewed. 

As shown in Fig. \ref{fig_1} (a), early "all-in-one" models such as \cite{chen2021pre}\cite{li2020all} proposed to train specialized headers and corresponding tails to handle each degradation of concern. Though model sharing the same backbone network, it is not convenient and lacks intelligence in practical applications. Especially, during training, it becomes an extremely complex and cumbersome procedure to train models for images with different levels and types of degradation. In inference phase, manual selection and identification of the corresponding inference model are laboriously required, too. 

To address this dilemma, some models\cite{li2022all}\cite{chen2022learning} proposed to directly and blindly restore degraded image with a single branch network in an end-to-end way, as shown in Fig. \ref{fig_1}(b). Since these models were unaware of the type of degradation they needed to handle, their performance were generally unsatisfactory when solving multiple restoration tasks. 

More recent models such as PromptIR\cite{potlapalli2023promptir}, CAPTNet\cite{gao2023prompt} and IDR\cite{zhang2023ingredient}, have been proposed to leverage learnable prompts to guide restoration, as shown in Fig. \ref{fig_1}(c). Nevertheless, they focused on learning visual prompts from distributional difference in training data. The existence of semantic gap\cite{pang2019towards} made it difficult for these models to accurately identify degradation types. Their restoration performances were therefore still just passable.

To address the above issues, we have proposed an effective textual prompt guided image restoration model named \textbf{TextPromptIR}, as shown in Fig. \ref{fig_1}(d). The learning process consisted of two stages. In the first stage, a task-specific BERT\cite{devlin2018bert} is fine-tuned to accurately understand textual instructions and generating semantic prompts for concerned all-in-one tasks. In the second stage, backbone network with different levels of depth-wise multi-head transposed attentions and gated convolution modules are proposed to fully utilize the semantic prompts obtained in the first stage. In the backbone, user-provided textual prompts integrated with visual information of image itself, enabled model accurate recognition of degradation type and guided model performing corresponding restoration operations.

\begin{figure}[b!]
\centering
\includegraphics[width=3in]{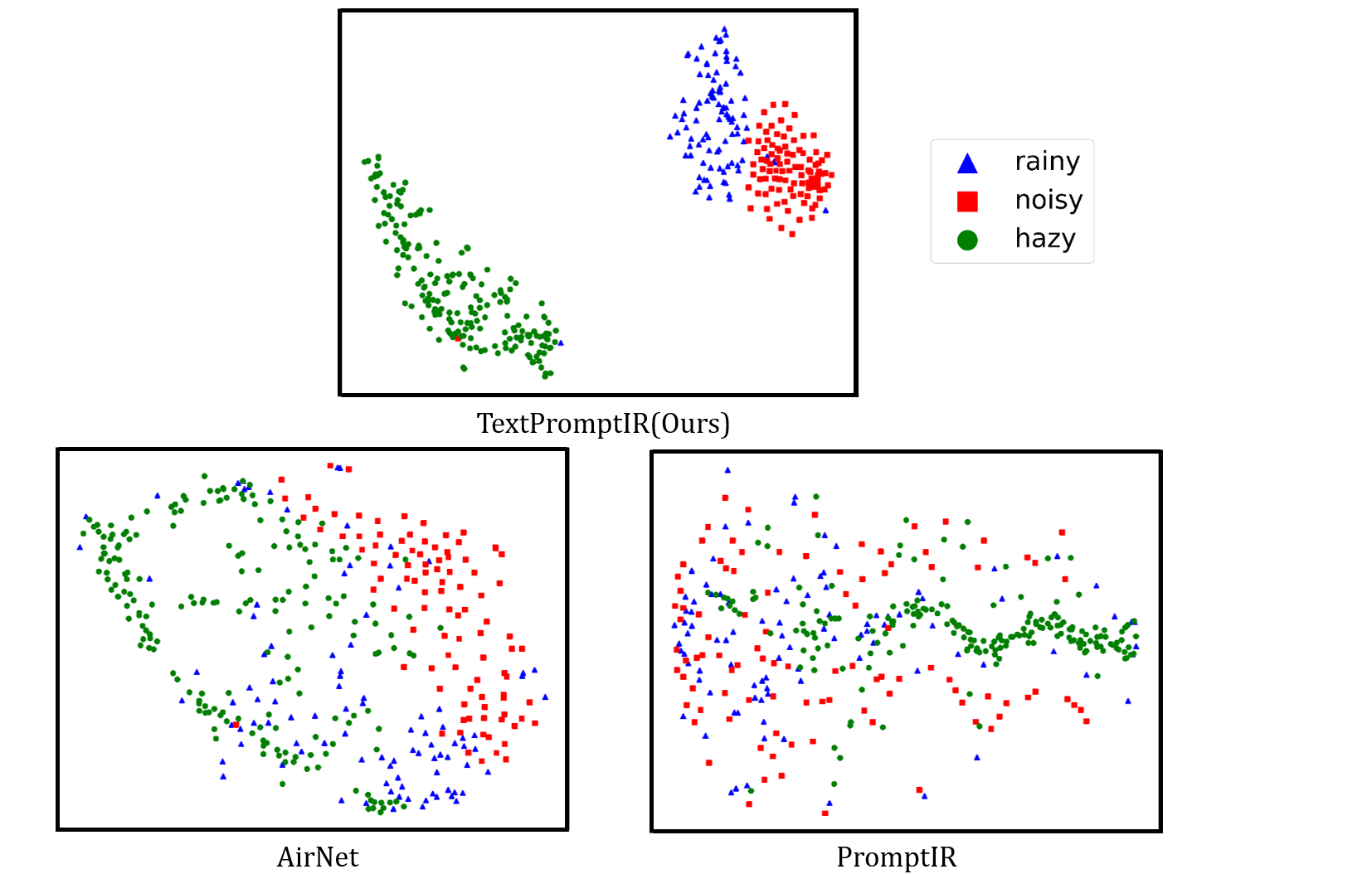}
\caption{t-SNE plots of degradation clustering identified by \textbf{TextPromptIR}, AirNet\cite{li2022all} and PromptIR\cite{potlapalli2023promptir}. Different colors denote different degradation type. Through semantic guidance, \textbf{TextPromptIR} is more capable of accurately identifying the degradation types and facilitating subsequent restoration.}
\label{fig_0}
\end{figure}

Compared to most of state-of-the-art methods\cite{li2022all,chen2022learning,potlapalli2023promptir,gao2023prompt,zhang2023ingredient}, \textbf{TextPromptIR} does not require cumbersome construction of different headers for different tasks. Additionally, it can effectively address semantic gap through fully utilizing textual prompting guidance, achieving precise restoration for images of different degradation type, as shown in Fig.\ref{fig_0}.

The main contributions are summarized as followings:
\begin{itemize}
\item We have proposed an effective textual prompt guided blind image restoration network. Through incorporating textual prompts, it can achieve accurate recognition and removal of various degradation in images without increasing the complexity of model. Extensive experiments on benchmarks demonstrate that the proposed method has achieved superior state-of-the-art performance in all-in-one image restoration tasks. 

\item We have specifically designed depth-wise multi-head transposed attentions and gated convolution modules to bridge the gap between textual prompts and visual features. The ablation studies have demonstrated their effectiveness.

\item We have innovatively introduced textual prompts into low-level visual domain and contributed an auxiliary multi-modal image restoration dataset. Our research highlights the potential of providing a natural, precise, and controllable interaction way for low-level image restoration tasks. It could provide meaningful references for future research on multimodal low-level vision.
\end{itemize}

\section{Related Works}
\subsection{Multi-degradation Image Restoration}
The objective of image restoration is to restore degraded images to their corresponding clear ones. Multi-degradation image restoration refers to a single unified model that can simultaneously handle multiple types of degradation tasks, also being referred to "all-in-one" image restoration task.
In this section, we focus on reviewing the recent developments on this research topic.

\subsubsection{Model with Independent Encoders and Decoders}
In early work, Li $et$ $al.$ \cite{li2020all} proposed an all-in-one model to address multiple degradation caused by unideal weather (e.g., rain, fog, snow). Each type of degradation is processed by a dedicated encoder and a universal decoder in their method. Chen $et$ $al.$ \cite{chen2021pre} constructed a transformer-based pretrained model to address multiple degradation, utilizing a structure composed of multi-heads and multi-tails. In inference phase, manual selections of the corresponding head and tail were also required.

The drawbacks of models in this category is obvious. As the number of tasks increased, the training process became extremely cumbersome and complex. Moreover, during inference, manually selecting modules for different degradation type was also a laborious and unnatural way.

\subsubsection{Single-branch End-to-End Model}
To address the above issues, many scholars have proposed to employ single branch end-to-end networks to solve multiple degradation problems. Typically, Li $et$ $al.$ \cite{li2022all} proposed to extract various degradation representations based on contrast learning, and utilize a single-branch network to address a wide range of degradation problems. In a similar vein, Chen $et$ $al.$ \cite{chen2022learning} achieved comparable results through learning an unified model across multiple restoration models based on knowledge distillation. 

However, the aforementioned models in this category cannot accurately identify the type of degradation. It limits these models to effectively handle inter-task dependencies. As a result, they cannot achieved satisfied restoration performance in complex scenes.

\subsubsection{Model with visual prompt}
With popular of prompt learning, researchers have proposed prompt-based image restoration models. Typically, Potlapalli $et$ $al.$ \cite{potlapalli2023promptir} and Gao $et$ $al.$ \cite{ gao2023prompt} encoded degradation-specific information and dynamically guided restoration network by utilizing learning prompts. Zhang $et$ $al.$ \cite{zhang2023ingredient} proposed a two-stage approach involving task-oriented knowledge collection and component-oriented knowledge integration. These models aimed to learn data distribution among degraded images and effectively train some adaptable prompt blocks for specific degradation. However, since persistent semantic gap \cite{pang2019towards} exists in training data, it hinders accurate identification of image degradation types. As a consequence, it remains a challenging pursuit on precise image restoration.

\subsection{Text-based prompt learning}
Prompt learning is originated from the field of natural language processing\cite{sarkar2019text}\cite{sood2020improving}. It aims to fully leverage the powerful content generation capabilities of large language models. Now, it has attracted wide attention, and become an useful learning mode in many research fields.  

Textual prompt learning has already played a crucial role in computer vision field. For example, Zhong $et$ $al.$ \cite{zhong2022regionclip} proposed a RegionCLIP model to align image regions with textual templates through feature matching. Shen $et al.$ \cite{shen2023text} utilized text-based question and answer pairs for object detection in videos. Bai $et al.$ \cite{bai2023TextIR} utilized prompts generated by CLIP model\cite{radford2021learning} to guide the process of image inpainting. Nichol $et$ $al.$ \cite{nichol2021glide}, Ramesh $et$ $al.$ \cite{ramesh2022hierarchical} and Zhou $et$ $al.$ \cite{zhou2023shifted} generated high-fidelity images through integrating textual information into diffusion model. 

Textual guidance can provide useful semantic information to help models better understand image, thereby improving the quality of final results. Moreover, textual guidance exhibits favorable characteristics of interactivity and engagement. Motivated by the advantages of textual prompt, in this work, we innovatively propose to introduce textual prompts into low-level visual domain, enriching multimodal research on low-level vision.

\section{The Proposed Method}
\subsection{Preliminaries}
In order to comprehend various types of multiple degradations, we briefly revisit the physical models of individual degradation such as noise, rain, haze, and so on. 

In the literature, the degradation process is commonly defined as following Equ.\ref{physic}:
\begin{equation}
\label{physic}
L = \phi(\lambda;H) + N ,
\end{equation}
where, $\phi(\cdot)$ denotes general degradation function, $\lambda$ denotes degradation parameters. $N$ represents additive noise, $L$ and $H$ respectively denote an observed low-quality image and its latent high-quality image. 

If $\phi(\cdot)$ represents element-wise addition, Equ.\ref{physic} can be reformulated to Equ.\ref{derain and denoise}:
\begin{equation}
\label{derain and denoise}
L = H + \lambda + N = H + \hat{\lambda} ,
\end{equation}
In many work, Equ.\ref{derain and denoise} is often employed to represent basic models of image deraining\cite{wang2022uformer} and image denoising\cite{dabov2007image}, where $\hat{\lambda}$ denotes rain streaks and \textit{i.i.d} zero-mean Gaussian noise, respectively. 

If $\phi(\cdot)$ represents element-wise multiplication, Equ.\ref{physic} can be reformulated to Equ.\ref{dehaze}:
\begin{equation}
\label{dehaze}
L = H \cdot \lambda + N = H \cdot \lambda + H \cdot \epsilon = H \cdot \hat{\lambda} ,
\end{equation}
In many work, Equ.\ref{dehaze} is then employed as the basic physic model of image dehazing\cite{song2023vision}, such as the atmosphere scattering model\cite{mccartney1976optics}\cite{narasimhan2002vision} and the Retinex theory\cite{land1977retinex}, where $\hat{\lambda}$ represents the transmission map or the atmospheric light.

All in all, it is apparent that different types of degradation task correspond to distinct formulations. These formulations effectively explain why a single image restoration model specifically designed for one degradation is inadequate to restore another degradation. Nonetheless, in real-world scenarios, diverse degradation coexists, necessitating an urgent demand of an unified model capable of simultaneously restoring multiple degradation.

As we know, the crucial aspect of an unified image restoration model for multi-degradation lies in its ability to accurately identify the degradation type, and dynamically adjust latent feature distributions for corresponding restoration operations. Therefore, it needs a natural, precise, and controllable interaction way to realize this goal. 

To address this issue, we are motivated to propose a textual prompt guided blind image restoration network. Semantic prompts from user help accurately identify the degradation type of image and interactively enhance the discriminative power of latent visual features in degradation space. 

We formulate the proposed network as a general function like Equ.\ref{TextPromptIR}:
\begin{equation}
\label{TextPromptIR}
\hat{I} = \textbf{TextPromptIR}(I,t),
\end{equation}
where, $I$ and $\hat{I}$ respectively represent a degraded image and its corresponding restored result. $t$ denotes user-provided textual prompts. 

\subsection{Network Architecture}
\begin{figure*}[htb]
\centering
\includegraphics[width=7in]{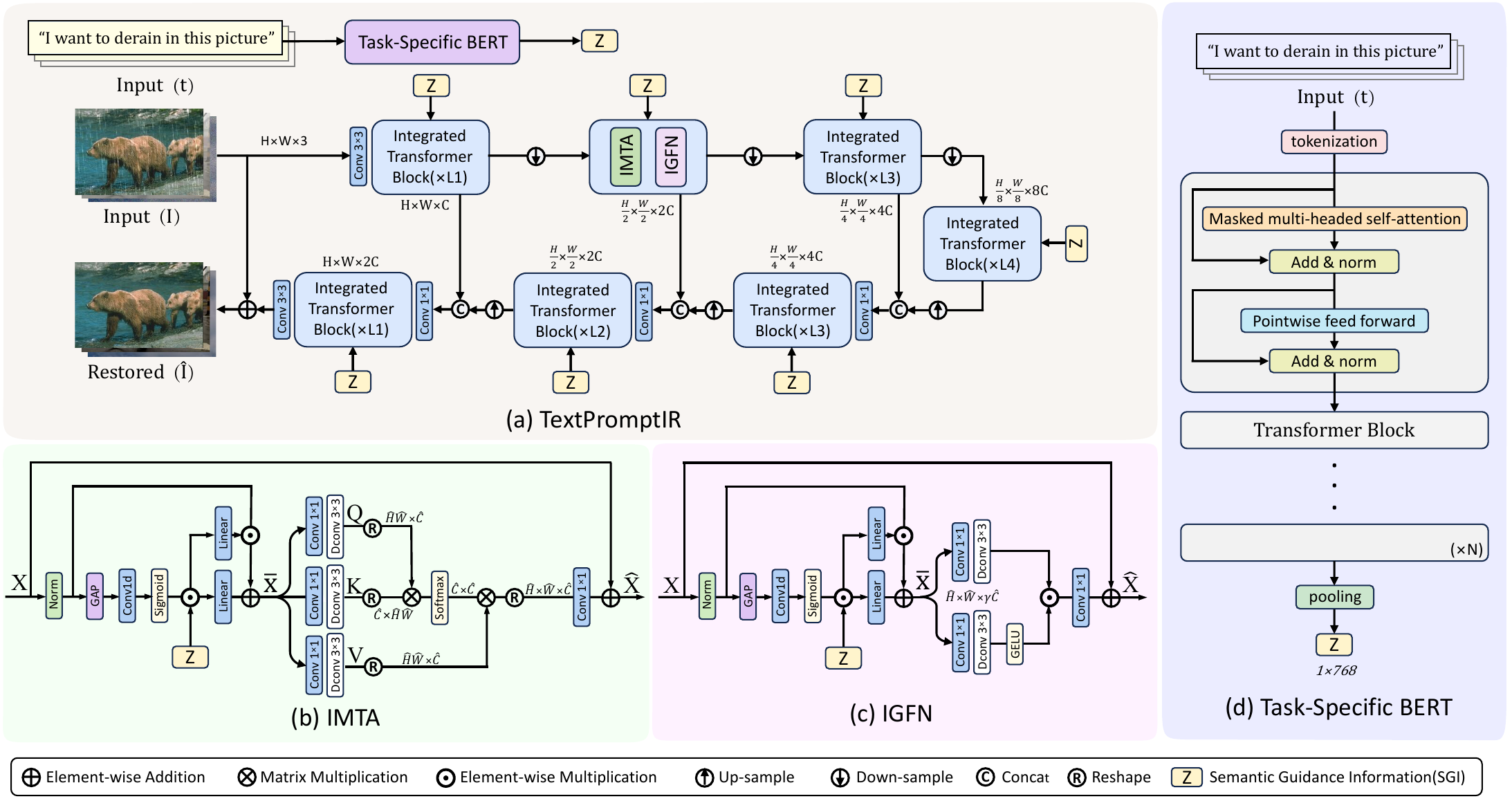}
\caption{(a) Overview of the proposed \textbf{TextPromptIR}. (b) Integrated Depth-wise Multi-head Transposed Attention (IMTA). (c) Integrated Depth-wise Gated Feed-Forward Network (IGFN). (d) Task-sepcific BERT.}
\label{fig_2}
\end{figure*}

The overview of the proposed network is shown in Fig. \ref{fig_2}. Its pipeline consists of two stage.

In the first stage, a task-specific BERT is fine-tuned by using our collected multi-modal image restoration dataset. Semantic guidance is then learned from user-provided textual instructions, by using the fine-tuned BERT. The learned semantic guidance is thereby taken as semantic prompts to control following learning process of backbone network. 

In the second stage, backbone network consisted of multi-level integrated transformer blocks (ITBlocks) is learned. These ITBlocks are stacked in an U-Net configuration. Each block is composed of an integrated depth-wise multi-head transposed attention (IMTA) and an integrated gated feed-forward convolution net (IGFN). 

In the following subsections, we will describe the overall pipeline and key components of the network in details. 

\subsubsection{Overall Pipeline}
Given an user-provided textual instruction $t$, a semantic guidance information $Z \in\mathbb{R}^{1 \times 768}$ is learned through task-specific BERT. 

Given a degraded image $I \in\mathbb{R}^{H \times W \times 3}$, a $3 \times 3$ convolution layer is applied to extract low-level feature mapping $F_0\in\mathbb{R}^{H \times W \times C}$ from $I$, where $H \times W$ represents the spatial dimensions and $C$ is the number of channel. 

The shallow feature $F_0$ is then passed through a multi-level encoder-decoder network, resulting in a high-level feature encoding $F_d\in\mathbb{R}^{H \times W \times 2C}$. Each $i^{th}$ encoder-decoder layer consists of $L_i$ ITBlocks. Semantic guidance information $Z$ is fed into each ITBlock as semantic prompt, helping accurately recognize degradation type and thereby guide corresponding restoration operations through cross-modal channel attention mechanism. 

Pixel-unshuffle and pixel-shuffle operations\cite{shi2016real} are respectively employed on feature upsampling and downsampling layers, for better preservation of feature information. Skip connections\cite{ronneberger2015u} between encoder-decoder layers at the same level are adopted in concatenation way for better feature reuse. After concatenation, a 1x1 convolution operation is performed to halve channels at respective level. This pipeline facilitates the preservation of fine structures and texture details during restoration process. 

Finally, a $3\times 3$ convolution layer is applied to the refined features to generate the predicted clear image $\mathbf{\hat I} \in\mathbb{R}^{H \times W \times 3}$

\subsubsection{Task-Specific BERT}
A task-specific BERT\cite{reimers2019sentence}\cite{devlin2018bert} for image restoration is utilized to generate semantic guidance.

Before fine-tuning the task-specific BERT, a multi-modal image restoration dataset is collected. With the help of the collected data, a pre-trained BERT is fine-tuned through Masked Language Model (MLM) and Next Sentence Prediction (NSP) tasks. As a result, semantic guidance information $\textbf{Z}\in\mathbb{R}^{1 \times 768}$ is obtained through average pooling word's BERT representations in sentence level.  

\subsubsection{Integrated Depth-wise Multi-head Transposed Attention}
To address the critical issue that how to effectively utilize semantic information to guide image restoration, we have proposed an integrated depth-wise multi-head transposed attention(IMTA) module. 

Specifically, given an image feature mapping $X\in\mathbb{R}^{H \times W \times C}$, a layer normalization\cite{ba2016layer} is first performed to obtain $X_0$. 

On one hand, for efficient fusion of image and semantic information, multi-modal feature map is learned. Specifically, a global average pooling is performed on $X_0$ in spatial dimension, resulting in $X_p\in\mathbb{R}^{1 \times C}$. Then a 1D convolution in channel direction and sigmoid nonlinear activation is utilized to obtain weight vector $\omega\in\mathbb{R}^{1 \times C}$. The semantic information $Z$ learned from fine-tuned BERT is linearly transformed and performs element-wise multiplication with the weight vector $\omega$, yielding multi-modal feature vector $X_w\in\mathbb{R}^{1 \times C}$. Then, $X_w$ is further fused with visual feature map $X_0$ through element-wise multiplication and addition in channels. Finally, a multi-modal feature map $X_c\in\mathbb{R}^{H \times W \times C}$ is obtained. The processing can be described in following Equ.\ref{IMTA}:
\begin{equation}
\label{IMTA}
\begin{gathered}
X_0 = LN(X),\\
\omega = \sigma(Conv_{1d}(GAP(X_0))),\\
X_w =  \omega \odot (W_l^1 Z),\\
X_c = (W_l^2 X_w) \odot X_0 + (W_l^3 X_w),
\end{gathered}
\end{equation}
where, $LN$ indicates layer normalization. $\odot$ represents element-wise multiplication. $Z$ is semantic guidance information. $Conv_{1d}$ and $GAP$ respectively refer to 1D convolution and global average pooling. $\sigma$ represents sigmoid activation. $W_l^1 Z$, $W_l^2 X_w$, $W_l^3 X_w$ $\in\mathbb{R}^{1 \times C}$.

On the other hand, to cope with high-resolution image scenarios, IMTA employs multi-head self-attention with linear complexity. Specifically, $X_c$ is projected into query $Q = W_d^Q W_s^Q X_c$, key $K = W_d^K W_s^K X_c$, and value $V = W_d^V W_s^V X_c$, where $W_s^{(\cdot)}$ is the $1 \times 1$ point-wise convolution and $W_d^{(\cdot)}$ is the $3 \times 3$ depth-wise convolution. Subsequently, dot product operations is performed between reshaped $\hat Q$ and $\hat K$ , generating transposed attention map $A \in \mathbb{R}^{\hat C \times \hat C}$. In this way, higher efficiency can be achieved, compared with conventional attention map of size $\mathbb{R}^{ H W \times H W}$\cite{dosovitskiy2020image}\cite{vaswani2017attention}.

Similar to traditional multi-head self-attention, we divide channels into multiple "heads", and learn separate attention maps in parallel, therefore, resulting in significant reduction in computational burden.

The process of aforementioned multi-head attention in channel direction can be described as following Equ.\ref{IMTA2}:
\begin{equation}
\label{IMTA2}
\hat X = W_s (\hat V \cdot Softmax(\hat K \cdot \hat Q / \beta)) + X,
\end{equation}
where, $\beta$ is a learnable scaling parameter that controls the magnitude of dot product between $\hat K$ and $\hat Q$. 

\subsection{Integrated Depth-wise Gated Feed-forward Convolution Net(IGFN)}
In order to accurately identify degradation types and utilize contextual information in image restoration, we propose an integrated gated feed-forward convolution net(IGFN). Multi-modal fusions and gate mechanism are employed to allow model to fully learn local image structures guided by semantic information. Complementing with IMTA, IGFN utilize contextual information to controls information flow for restoration. The formulation is as shown in following Equ.\ref{IGFN}:
\begin{equation}
\begin{gathered}
\label{IGFN}
\hat X = W_s^0 Gate(X_c) + X_c,\\
Gate(X_c) = \phi(W_d^1 W_s^1 X_c)\odot (W_d^2 W_s^2 X_c),
\end{gathered}
\end{equation}
where, the process to obtain $X_c$ is similar to Equ.\ref{IMTA}, $\odot$ indicates element-wise multiplication, $\phi$ is GELU activation\cite{hendrycks2016gaussian}. 

\section{Experiments}
In order to demonstrate the effectiveness of our proposed \textbf{TextPromptIR}, we have conducted extensive experiments on various datasets. In this section, we will describe experiment settings and present qualitative and quantitative results for analysis. Finally, we perform ablation studies to discuss the impacts of each proposed key components.

\subsection{Experiment Settings}
\subsubsection{Datasets}
BSD400\cite{martin2001database}, CBSD68\cite{martin2001database}, WED\cite{ma2016waterloo}, and Urban100\cite{huang2015single} are employed for image denoising, Rain100L\cite{yang2017deep} for image deraining, and RESIDE\cite{li2018benchmarking} for image dehazing. 

In terms of denoising task, BSD400 dataset consists of 400 clear natural images, CBSD68 comprises 68 clear color images, Urban100 contains 100 clear natural images, and WED encompasses a total of 4744 images. WED and BSD400 are jointly taken as training set. Urban100 and CBSD68 are utilized as testing sets. As the same settings as in \cite{tian2020image,zhang2017beyond,zhang2017learning,zhang2018ffdnet}, three levels of Gaussian noise $\sigma = \{15,25,50\}$ are respectively added to clear images, generating corresponding noisy images for training and evaluation. 

In terms of deraining task, Rain100L\cite{yang2017deep} includes 200 rainy-clear training image pairs and 100 testing image pairs. 

In terms of dehazing task, within RESIDE\cite{li2018benchmarking}, SOTS dataset with 72,135 images is taken as training set. OTS dataset with 500 images is for testing. 

To facilitate our textual prompt learning, we have contributed an auxiliary multimodal dataset. Specifically, the dataset is augmented by using prompts like "generate with [input text] with similar meanings." on GPT3.5\cite{brown2020language}. 150 sentences are collected in total. These 150 sentences are taken as semantic instruction dataset paired with the aforementioned image restoration datasets. They are categorized into distinct deraining, dehazing, and denoising categories. Each category includes 50 sentences with different description while having similar semantic. During both training and testing phases, a sentence is randomly selected as a paired semantic instruction for image from corresponding category. As a result, a multimodal dataset is dynamically created for model learning.

\subsubsection{Implementation Details}
The numbers of ITBlocks in the proposed \textbf{TextPromptIR} are respectively set [4,6,6,8] ranging from Level1 to Level4. Their channel numbers are respective [48,96,192,384]. The attention heads are correspondingly set as [1,2,4,8]. 

All experiments are conducted using PyTorch on a NVIDIA GeForce RTX 3090 GPU. Adam optimizer\cite{kingma2014adam} with parameters ($\beta_1=0.9$, $\beta_2=0.999$, and weight decay $1\times10^{-4}$) are employed. The initial learning rate is set to be $2\times10^{-4}$. During training, the batch size is 6. Additionally, random horizontal and vertical flips are applied to images for data augmentation. Images are cropped in patch size of $128\times128$ for training.

\subsubsection{Evaluation metrics}
Following previous work \cite{dong2020fd,fu2019lightweight,tian2020image}, we employ Peak Signal-to-Noise Ratio (PSNR)\cite{huynh2008scope} and Structural Similarity (SSIM)\cite{wang2004image} as our quantitative evaluation metrics. In all performance tables, the best and second-best performances are denoted with bold and underlined annotations, respectively.

\subsection{Comparisons on Multiple Degradation}
One appealing aspect of \textbf{TextPromptIR} is its incorporation of semantic prompt to guide image restoration in all-in-one task. To demonstrate it, we compare the proposed method with several popular state-of-the-art methods. We select four methods on image restoration for single degradation (IRSD) ($i.e.$, BRDNet\cite{tian2020image}, LPNet\cite{fu2019lightweight}, FDGAN\cite{dong2020fd} and MPRNet\cite{zamir2021multi}) and five methods on image restoration for multiple degradation (IRMD) ($i.e.$, TKMANet\cite{chen2022learning}, DL\cite{fan2019general}, AirNet\cite{li2022all}, DA-CLIP\cite{luo2023controlling} and PromptIR\cite{potlapalli2023promptir}). 

To ensure a fair and accurate comparison, the training and testing settings are kept the same as the ones in the compared methods. From Table \ref{tab:table1}, we can observe that \textbf{TextPromptIR} outperforms almost all models. It indicates that the proposed model can accurately restore various degradation types with semantic guidance. 

\begin{table*}[t]
\renewcommand\arraystretch{1}
\caption{Comparisons with state-of-the-art all-in-one image restoration methods under all-in-one restoration setting.\label{tab:table1}}
\centering
\begin{tabular}{c|ccccc|c}
   \hline
   \multirow{2}{*}{\centering Method} & \multicolumn{3}{c}{Denoising on CBSD68 dataset} & Deraining & Dehazing & \multirow{2}{*}{\centering Average}\\
   & $\sigma$ = 15 & $\sigma$ = 25 & $\sigma$ = 50 & on Rain100L & on SOTS & \\
   \hline
   BRDNet\cite{tian2020image} & 32.26/0.898 & 29.76/0.836 & 26.34/0.693 & 27.42/0.895 & 23.23/0.895 & 27.80/0.843\\
   LPNet\cite{fu2019lightweight} & 26.47/0.778 & 24.77/0.748 & 21.26/0.552 & 24.88/0.784 & 20.84/0.828 & 23.64/0.738\\
   FDGAN\cite{dong2020fd} & 30.25/0.910 & 28.81/0.868 & 26.43/0.776 & 29.89/0.933 & 24.71/0.929 & 28.02/0.883\\
   MPRNet\cite{zamir2021multi} & 33.54/0.927 & 30.89/0.880 & 27.56/0.779 & 33.57/0.954 & 25.28/0.955 & 30.17/0.899\\
   DL\cite{fan2019general} & 33.05/0.914 & 30.41/0.861 & 26.90/0.740 & 32.62/0.931 & 26.92/0.931 & 29.98/0.875\\
   AirNet\cite{li2022all} & 33.92/\underline{0.933} & 31.26/\underline{0.888} & 28.00/0.797 & 34.90/0.968 & 27.94/0.962 & 31.20/0.910\\
   TKMANet\cite{chen2022learning} & 33.02/0.924 & 30.31/0.820 & 23.80/0.556 & 34.94/0.972 & 30.41/0.973 & 30.50/0.849\\
   DA-CLIP\cite{luo2023controlling} & 30.02/0.821 & 24.86/0.585 & 22.29/0.476 & 36.28/0.968 & 29.46/0.963 & 28.58/0.763 \\
   PromptIR\cite{potlapalli2023promptir} & \underline{33.98}/\underline{0.933} & \underline{31.31}/\underline{0.888} & \underline{28.06}/\underline{0.799} & \underline{36.37}/\underline{0.972} & \underline{30.58}/\underline{0.974} & \underline{32.06}/\underline{0.913}\\
   \hline
   TextPromptIR(Ours) & \textbf{34.17}/\textbf{0.936} & \textbf{31.52}/\textbf{0.893} & \textbf{28.26}/\textbf{0.805} & \textbf{38.41}/\textbf{0.982} & \textbf{31.65}/\textbf{0.978} & \textbf{32.80}/\textbf{0.919}\\
   \hline
\end{tabular}
\end{table*}

\begin{figure*}[!h]
\centering
\includegraphics[width=7in]{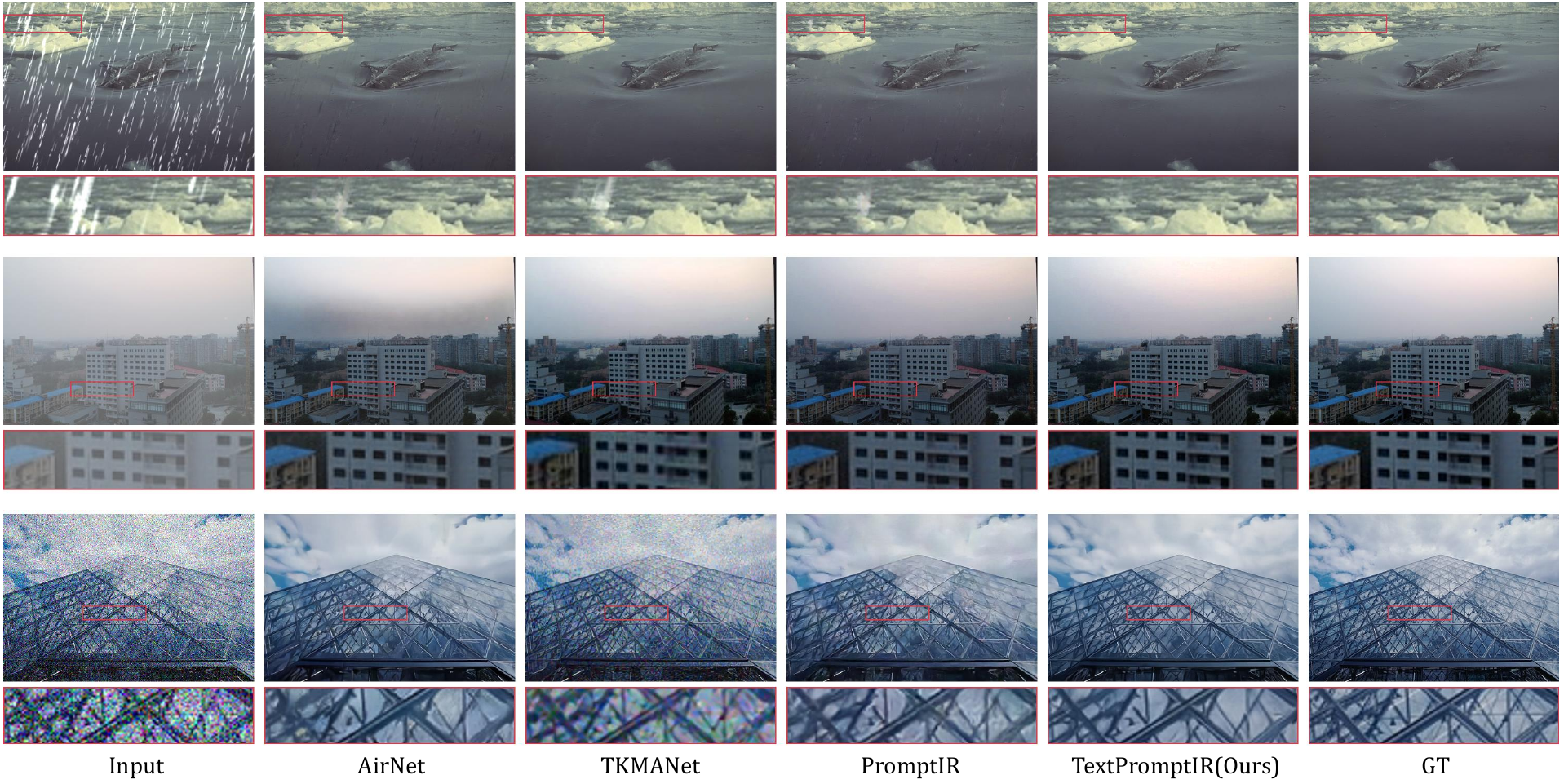}
\caption{Visual comparisons with SOTA all-in-one models on Rain100L\cite{yang2017deep}, SOTS\cite{li2018benchmarking} and CBSD68\cite{martin2001database} sample images. The proposed model exhibits precise degradation removal and better preservation of image content details.}
\label{fig_3}
\end{figure*}

The visual results in Fig. \ref{fig_3} demonstrate that \textbf{TextPromptIR} can accurately remove different types of degradation while preserving more details.

\subsection{Comparisons on Single Degradation}
In this section, we primarily evaluate the performance of \textbf{TextPromptIR} under single-task mode. Individual model is trained for a corresponding restoration task. Experimental results show that \textbf{TextPromptIR} not only achieves significant improvements in the all-in-one task, but also competes favorably with state-of-the-art methods in single-task mode. Specifically, From Table \ref{tab:table2}, we can observe that \textbf{TextPromptIR} achieves an average 0.13dB PSNR higher than the second-best method Restormer\cite{zamir2022restormer}, on the Urban100, Rain100L and SOTS datasets.

\begin{figure*}[ht]
\centering
\includegraphics[width=7in]{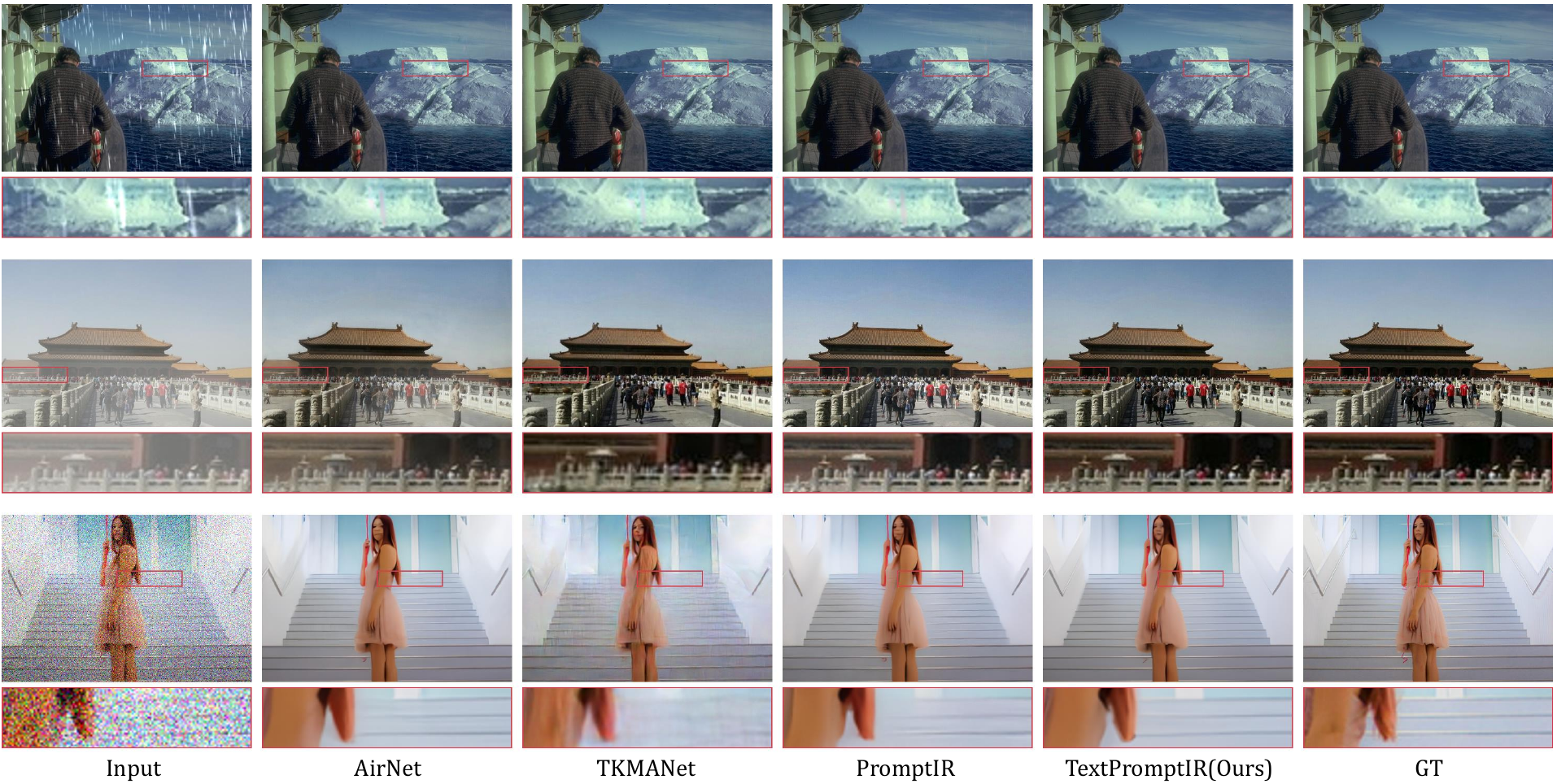}
\caption{Visual comparisons with SOTA models under single-task conditions on Rain100L\cite{yang2017deep}, SOTS\cite{li2018benchmarking} and Urban100\cite{huang2015single} sample images. The proposed model can accurately restore images while preserving well image details.}
\label{fig_4}
\end{figure*}

As shown in Fig. \ref{fig_4}, the visual results demonstrate that \textbf{TextPromptIR} is capable of preserving more details. Our restored images align better with human perceptual quality.

\begin{table*}[t]
\renewcommand\arraystretch{1}
\caption{Comparisons with state-of-the-art image restoration methods under one-by-one restoration setting. \label{tab:table2}}
\centering
\begin{tabular}{c|ccccc|c}
   \hline
   \multirow{2}{*}{\centering Method} & \multicolumn{3}{c}{Denoising on Urban100 dataset} & Deraining & Dehazing & \multirow{2}{*}{\centering Average}\\
   & $\sigma$ = 15 & $\sigma$ = 25 & $\sigma$ = 50 & on Rain100L & on SOTS & \\
   \hline
   MPRnet\cite{zamir2021multi} & 34.55/0.949 & 32.26/0.925 & 29.06/0.873 & 38.26/0.982 & 28.21/0.967 & 32.47/0.939\\
   AirNet\cite{li2022all} & 34.40/0.949 & 32.10/0.924 & 28.88/0.870 & 34.90/0.966 & 23.18/0.900 & 30.69/0.922\\
   Restormer\cite{zamir2022restormer} & 34.67/\textbf{0.969} & 32.41/0.927 & 29.31/0.878 & \underline{38.99}/0.978 & 30.87/0.969 & \underline{33.25}/\underline{0.944}\\
   TKMANet\cite{chen2022learning} & 33.84/0.941 & 30.34/0.860 & 26.86/0.780 & 35.60/0.974 & \underline{31.37}/\underline{0.974} & 31.60/0.906\\
   PromptIR\cite{potlapalli2023promptir} & \textbf{34.77}/\underline{0.952} & \underline{32.49}/\textbf{0.929} & \underline{29.39}/\underline{0.881} & 37.04/\underline{0.979} & 31.31/0.973 & 33.00/0.943\\
   \hline
   TextPromptIR(Ours) & \underline{34.76}/0.951 & \textbf{32.54}/\textbf{0.929} & \textbf{29.47}/\textbf{0.882} & \textbf{39.03}/\textbf{0.985} & \textbf{31.74}/\textbf{0.981} & \textbf{33.51}/\textbf{0.946}\\
   \hline
\end{tabular}
\end{table*}

\subsection{Experiments in Real-World Scenarios}
\begin{figure*}[htb]
\centering
\includegraphics[width=7in]{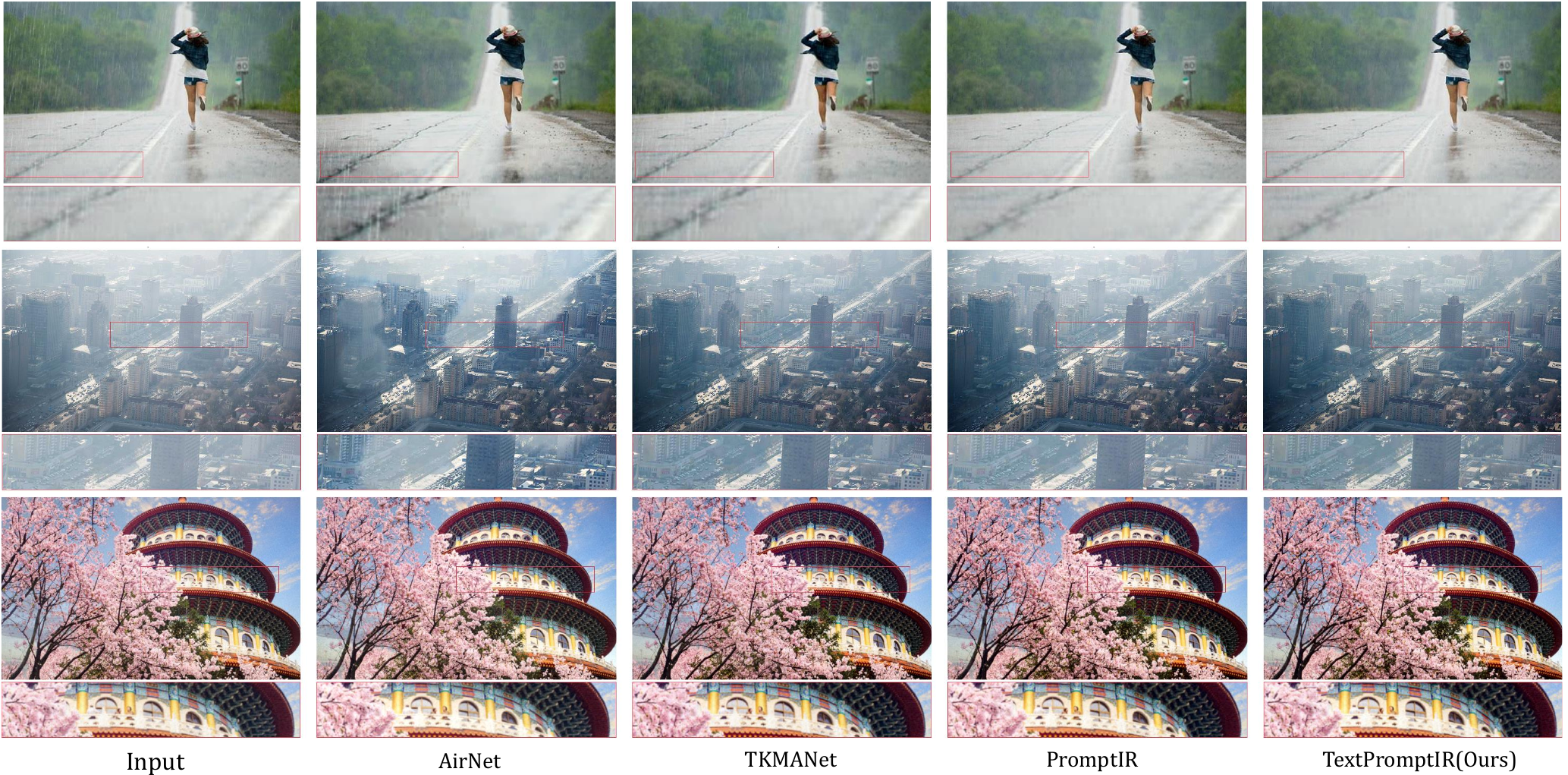}
\caption{Visual comparisons on real-world degraded scenes.}
\label{fig_7}
\end{figure*}

The proposed \textbf{TextPromptIR} is characterized that it can effectively bridge semantic gap inherent in image. To demonstrate this point, we have selected three representative images from real-world scenarios, and performed corresponding deraining/dehazing/denoising tasks. As shown in Fig. \ref{fig_7}, the resulted images reveal that \textbf{TextPromptIR} exhibits more pronounced removal results compared to other all-in-one models' (e.g. AirNet\cite{li2022all}, PromptIR\cite{potlapalli2023promptir}, TKMANet\cite{chen2022learning}) and preserves better image's structural details.

\subsection{Confirmatory Experiments: Model's Comprehension on Semantic Prompts}
One compelling aspect of \textbf{TextPromptIR} is its ability to perform corresponding tasks based on human intent within an unified framework. To validate the impacts of user's instructions, we conducted two confirmatory experiments. The first one is for single degradation. We provide model with unrelated textual prompts. The second one is for images containing multiple complex degradation. We likewise utilized different textual prompts.

\subsubsection{Image with Single Degradation} 
\begin{figure*}[htb]
\centering
\includegraphics[width=7in]{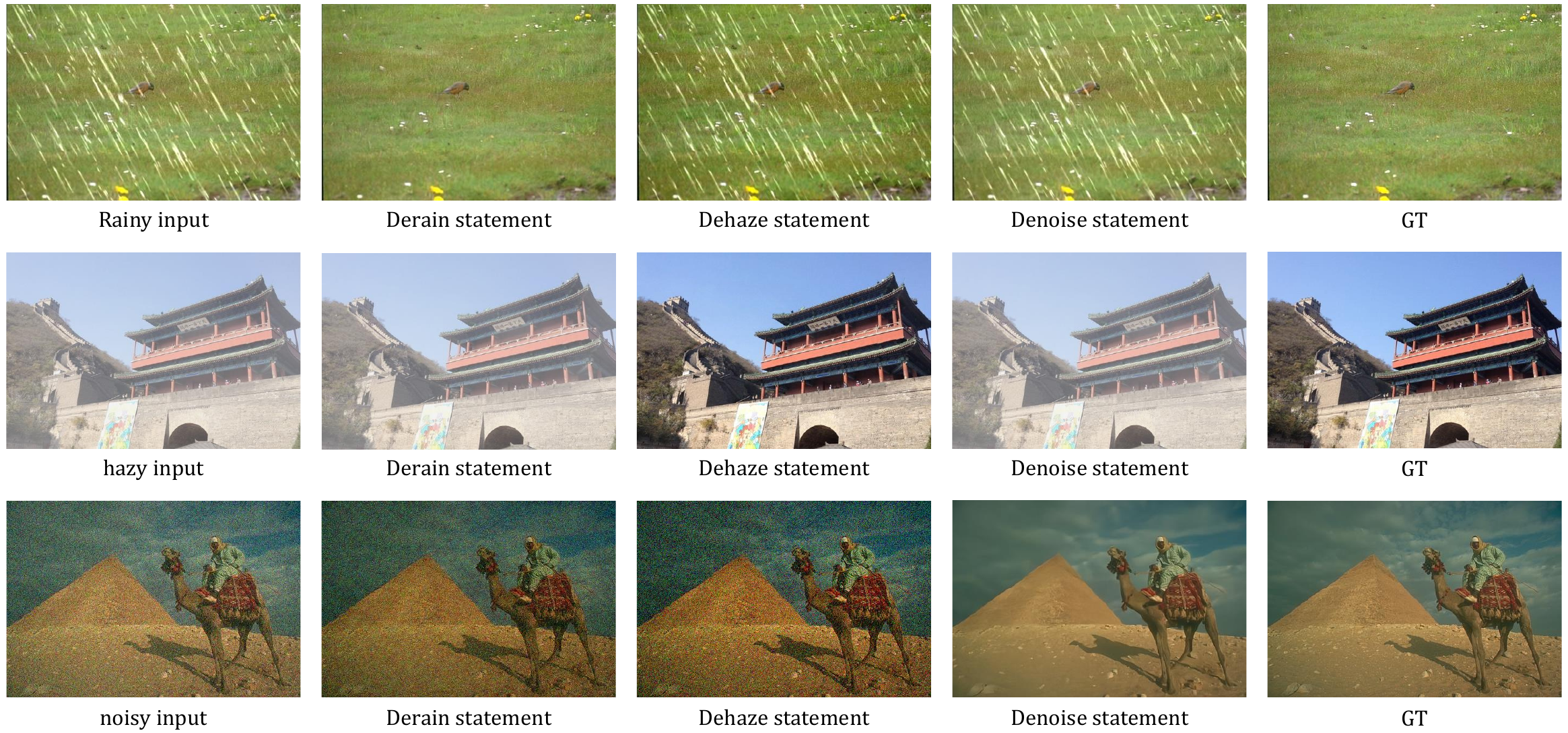}
\caption{Confirmatory experiment I: various textual prompts for the same image with single degradation.}
\label{fig_5}
\end{figure*}

As shown in Fig. \ref{fig_5}, we provide the same degraded image with various textual prompts. We observed that right \textbf{TextPromptIR} can effectively facilitate restoration, if it receives right prompt that is aligned with corresponding degradation. On contrast, when receiving unrelated prompts, it prefers minimal changes to the degradation.

\subsubsection{Image with Multiple Complex Degradations} 
\begin{figure*}[!h]
\centering
\includegraphics[width=7in]{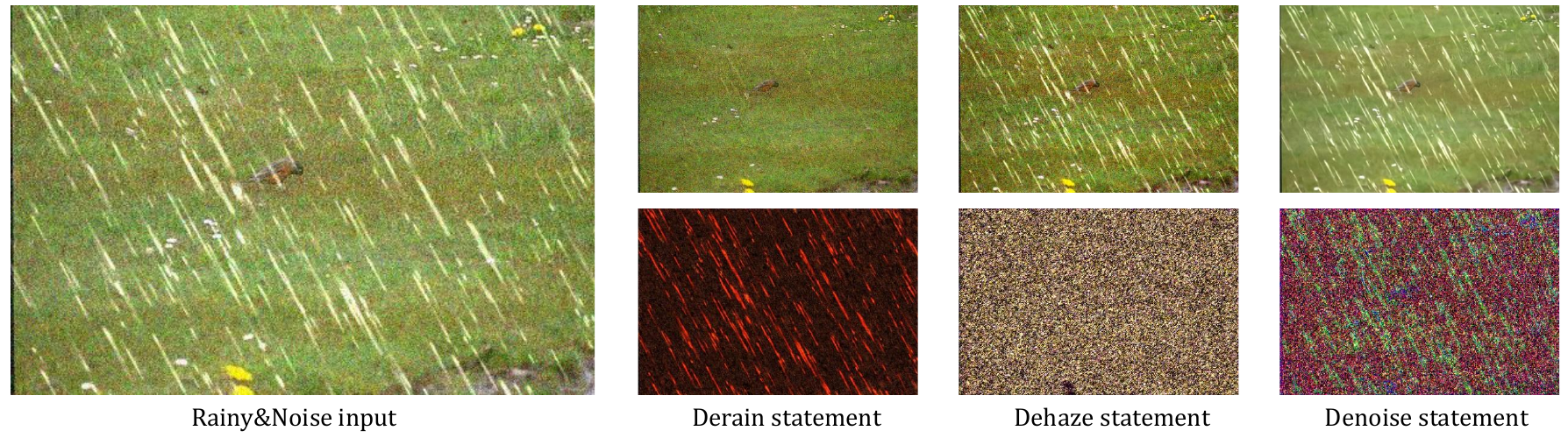}
\caption{Confirmatory experiment II: various textual prompts for the same image with multiple degradation, along with corresponding attention maps.}
\label{fig_6}
\end{figure*}

As shown in Fig. \ref{fig_6}, in a coupled scenario with multiple degradations, we can observe that textual prompts with distinct task enable precise removal of corresponding degradation while unchanging the unrelated degradation. 

Furthermore, we incorporate attention layers to visualize the inherent mechanism. Under the guidance of semantic information, the model accurately comprehended the types of degradation and achieved precise restoration. This demonstrates that the proposed model can well accomplish tasks in accordance with human intent.

\subsection{Ablation Studies}
\subsubsection{Impacts of Degradation Combination on Model Training}
In this section, we analyze the impact of different degradation combinations on the performance of the proposed \textbf{TextPromptIR} model.
As shown in Table \ref{tab:table6}, It is interesting discover that joint training for denoising and deraining can significantly improve the performance of deraining. However, when dehazing is jointly trained with deraining or denoising, the performances somewhat decrease compared to the case that all three degradations are jointly trained. The observation constitutes an intriguing problem worth further exploration, which is out of scope in this work.
\begin{table*}[htb]
\renewcommand\arraystretch{1}
\caption{Ablation study on degradation combination. "$\checkmark$" represents \textbf{TextPromptIR} for corresponding degradation combination case, "-" denotes unavailable results. The best results are highlighted in \textbf{bold}.}
\label{tab:table6}
\centering
\begin{tabular}{ccc|ccc|c|c}
   \hline
   \multicolumn{3}{c|}{Degration} & \multicolumn{3}{c|}{Denoise} & Derain & Dehaze\\
   Noise & Rain & Haze & CBSD68($\sigma$ = 15) & CBSD68($\sigma$ = 25) & CBSD68($\sigma$ = 50) & Rain100L & SOTS\\
   \hline
    $\checkmark$ &  &  & \textbf{34.33}/\textbf{0.938} & \textbf{31.69}/\textbf{0.897} & \textbf{28.45}/\textbf{0.813} & - & - \\
     & $\checkmark$ &  & - & - & - & 39.03/\textbf{0.985} & - \\
     &  & $\checkmark$ & - & - & - & - & \textbf{31.74}/\textbf{0.981} \\
    $\checkmark$ & $\checkmark$ &  & 34.20/0.936 & 31.56/0.893 & 28.32/0.806 & \textbf{39.20}/\textbf{0.985} & -\\
    $\checkmark$ &  & $\checkmark$ & 34.07/0.933 & 31.52/0.893 & 28.29/0.806 & - & 31.24/0.977 \\
     & $\checkmark$ & $\checkmark$ & - & - & - & 38.35/0.983 & 31.24/0.979 \\
    $\checkmark$ & $\checkmark$ & $\checkmark$ & 34.17/0.936 & 31.52/0.893 & 28.26/0.805 & 38.41/0.982 & 31.65/0.978 \\
   \hline
\end{tabular}
\end{table*}

\subsubsection{Impacts of Semantic Guidance by Textual Prompt}
We conducted experiments to validate the effectiveness of semantic guidance. To do so, we removed semantic guidance module from backbone network. The ablation performances are shown in Table \ref{tab:table7}. It can be observed that the performance of ablation model significantly declined. It indicates that semantic guidance information plays a paramount role in recognizing and removing degradation in all-in-one task.

\begin{table*}[htb]
\renewcommand\arraystretch{1}
\caption{Ablation study on the impact of semantic guidance information (SGI). The best results are highlighted in \textbf{bold}.}
\label{tab:table7}
\centering
\begin{tabular}{c|ccc|c|c}
   \hline
   \multirow{2}{*}{\centering Dataset} & \multicolumn{3}{c|}{Denoise} & Derain & Dehaze\\
   & CBSD68($\sigma$ = 15) & CBSD68($\sigma$ = 25) & CBSD68($\sigma$ = 50) & Rain100L & SOTS\\
   \hline
    w.o. SGI & 33.97/0.933 & 31.33/0.889 & 28.02/0.798 & 37.91/0.981 & 30.75/0.975 \\
    Ours & \textbf{34.17}/\textbf{0.936} & \textbf{31.52}/\textbf{0.893} & \textbf{28.26}/\textbf{0.805} & \textbf{38.41}/\textbf{0.982} & \textbf{31.65}/\textbf{0.978} \\
   \hline
\end{tabular}
\end{table*}

\subsubsection{Model Stability}
\begin{figure}[htb]
\centering
\includegraphics[width=2.5in]{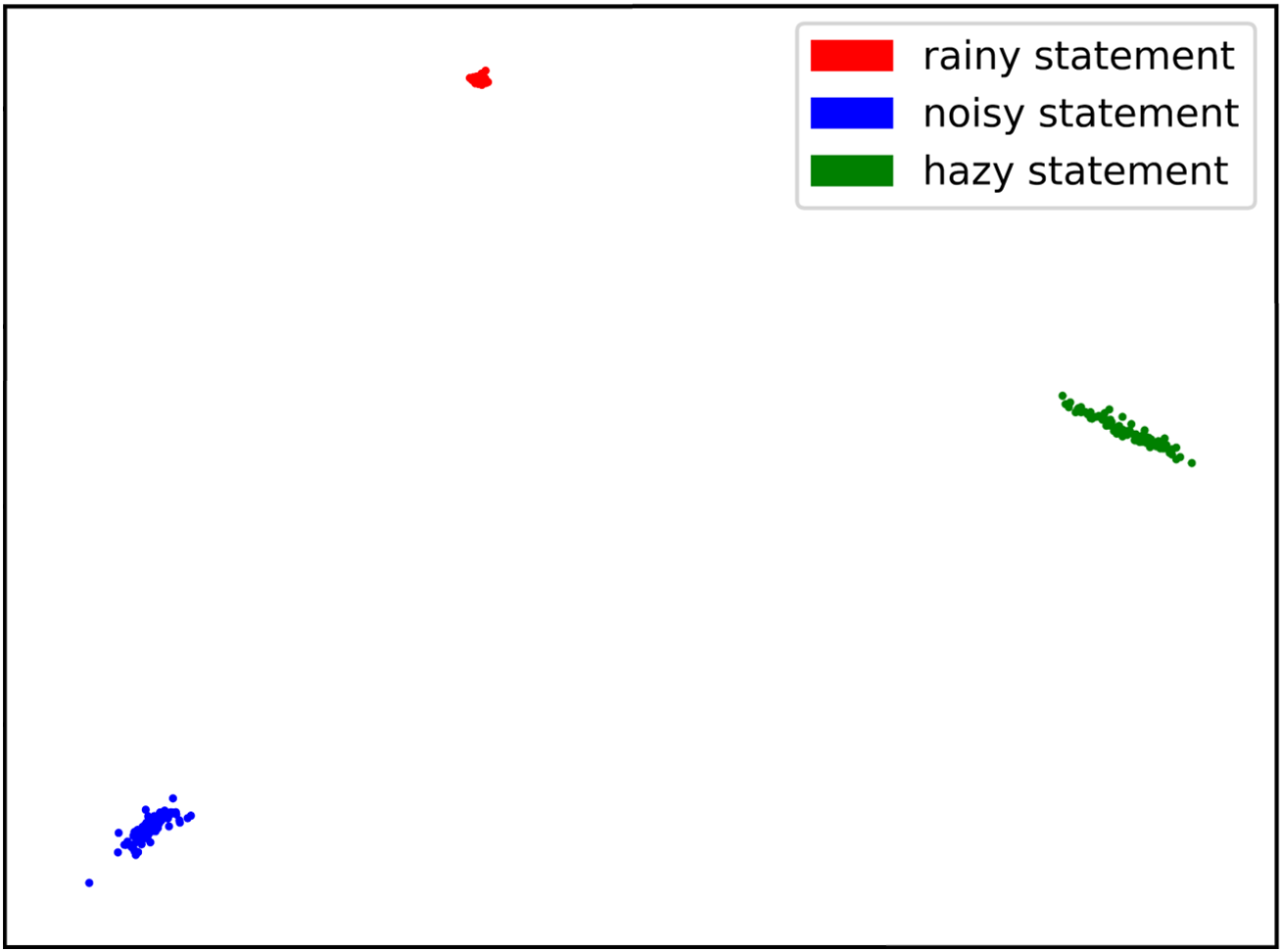}
\caption{t-SNE plots of guidance features learned from the task-specific BERT. The semantic prompt on user's intent can be accurately recognized.}
\label{fig_8}
\end{figure}

In this section, we investigate the stability of the proposed model. As shown in Fig. \ref{fig_8}, the task-specific BERT exhibits precise recognition capability. It can accurately extract the semantic information of user-provided prompt sentences. 

\begin{figure}[htb]
\centering
\includegraphics[width=3.5in]{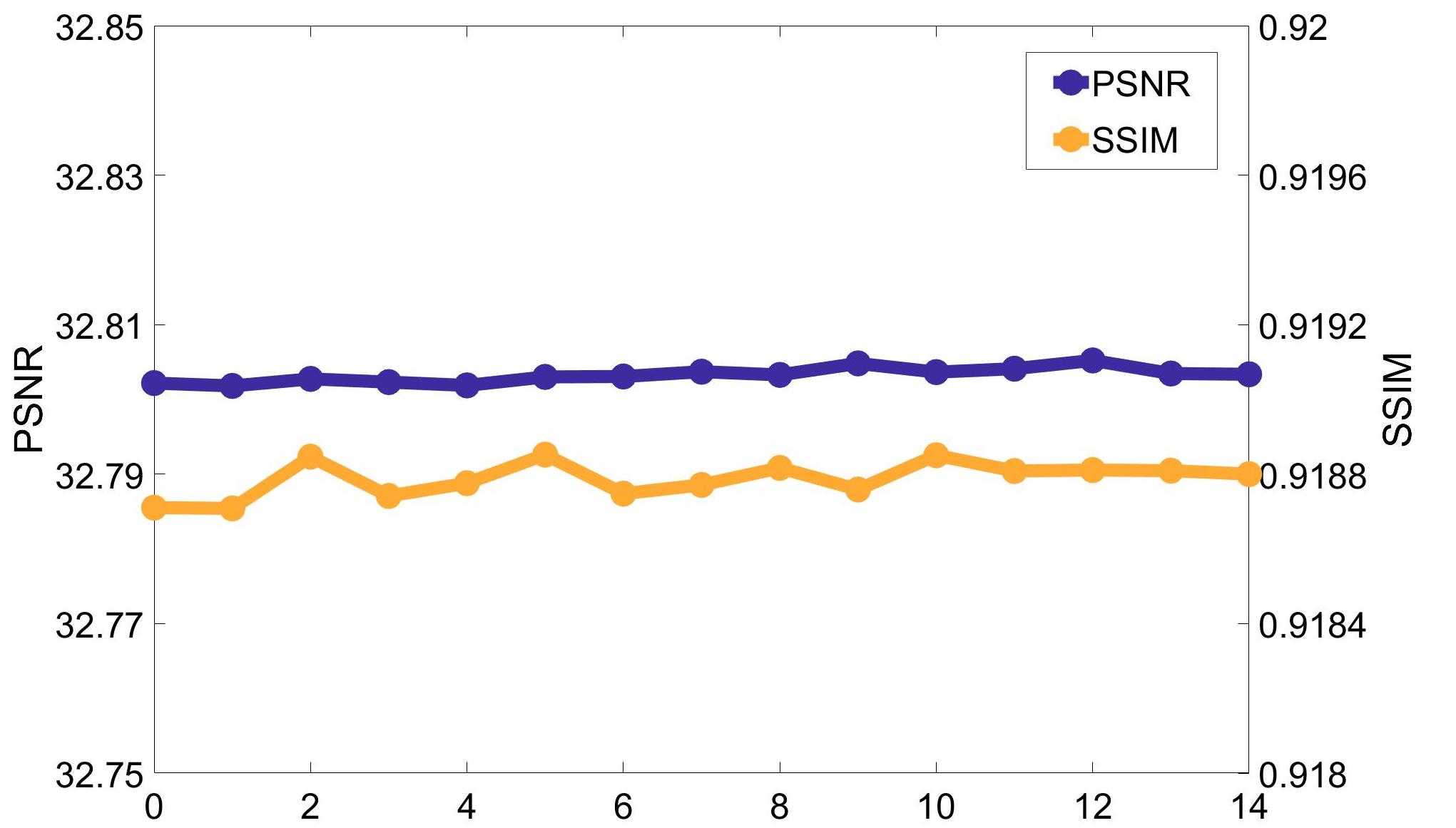}
\caption{The average PSNR and SSIM of 15 repeated experiments. It indicates the high stability of model when facing different textual prompts of similar task instruction.}
\label{fig_9}
\end{figure}

We further conduct fifteen repeated experiments using random seeds. In each experiment, one right textual prompt is randomly selected, and employed as restoration guidance for the same degraded image. As illustrated in Fig. \ref{fig_9}, though the learned representations from task-specific BERT are not exactly consistency, they have little impacts on model's performance. This observation indicates that the proposed model is robust to prompt sentences with similar semantic.

\section{Conclusion}
In this paper, we have proposed an effective textual prompt guided all-in-one image restoration model. Through incorporating semantic prompts, the proposed model can achieve accurate recognition and removal of various image degradation without increasing model's complexity. Extensive experiments on public benchmarks have demonstrated that the proposed model can achieve much more superior state-of-the-art performance in denoising/deraining/dehazing tasks. The proposed model provides a natural, precise, and controllable interaction way for future low-level image restoration research. 


\section*{Acknowledgement}
This work is supported by National Natural Science Foundation of China under Grand No. 62366021.

\bibliographystyle{elsarticle-num}  
\bibliography{references}  

\end{document}